\documentclass[sigconf]{acmart}

\AtBeginDocument{%
  }

\copyrightyear{2025}
\acmYear{2025}
\setcopyright{cc}
\setcctype{by}\setcctype{by}
\acmConference[ICMI Companion '25]{Companion Proceedings of the 27th
International Conference on Multimodal Interaction}{October 13--17,
2025}{Canberra, ACT, Australia}
\acmBooktitle{Companion Proceedings of the 27th International Conference on
Multimodal Interaction (ICMI Companion '25), October 13--17, 2025, Canberra,
ACT, Australia}
\acmDOI{10.1145/3747327.3764793}
\acmISBN{979-8-4007-2076-5/2025/10}

\usepackage[inline]{enumitem}
\usepackage{algorithmic}
\usepackage{graphicx}
\usepackage{textcomp}
\usepackage{xcolor}

\usepackage{witharrows}

\usepackage[flushleft]{threeparttable}
\usepackage{tablefootnote}
\usepackage{array}
\newcolumntype{P}[1]{>{\centering\arraybackslash}p{#1}}
\usepackage{multicol}
\usepackage{multirow}
\usepackage{tabulary}
\usepackage{colortbl}
\definecolor{mygray}{gray}{0.90}
\usepackage{makecell}
\usepackage{booktabs}
\usepackage{subcaption}

\usepackage{stmaryrd}

\usepackage{float}
\usepackage{pifont}

\usepackage{arydshln}
\usepackage[switch,columnwise]{lineno}

\begin{document}

\title{Multi-Representation Diagrams for Pain Recognition: 
Integrating Various Electrodermal Activity Signals into a Single Image }

\author{Stefanos Gkikas}
\email{gkikas@ics.forth.gr}
\orcid{0000-0002-4123-1302}
\affiliation{%
  \institution{Foundation for Research \& Technology-Hellas}
  \city{Heraklion}
  \country{Greece}
}

\author{Ioannis Kyprakis }
\email{ikyprakis@ics.forth.gr}
\orcid{0009-0000-0711-5108}
\affiliation{%
  \institution{Foundation for Research \& Technology-Hellas}
  \city{Heraklion}
  \country{Greece}
}

\author{Manolis Tsiknakis}
\email{tsiknaki@ics.forth.gr}
\orcid{0000-0001-8454-1450}
\affiliation{%
  \institution{Foundation for Research \& Technology-Hellas and Hellenic Mediterranean University}
  \city{Heraklion}
  \country{Greece}
}


\begin{abstract}
Pain is a multifaceted phenomenon that affects a substantial portion of the population. Reliable and consistent evaluation benefits those experiencing pain and underpins the development of effective and advanced management strategies.
Automatic pain-assessment systems deliver continuous monitoring, inform clinical decision-making, and aim to reduce distress while preventing functional decline. By incorporating physiological signals, these systems provide objective, accurate insights into an individual's condition.
This study has been submitted to the \textit{Second Multimodal Sensing Grand Challenge for Next-Gen Pain Assessment (AI4PAIN)}. 
The proposed method introduces a pipeline that leverages electrodermal activity signals as input modality. Multiple representations of the signal are created and visualized as waveforms, and they are jointly visualized within a single multi-representation diagram. 
Extensive experiments incorporating various processing and filtering techniques, along with multiple representation combinations, demonstrate the effectiveness of the proposed approach. It consistently yields comparable, and in several cases superior, results to traditional fusion methods, establishing it as a robust alternative for integrating different signal representations or modalities.

\end{abstract}




\maketitle

\section{Introduction}
Pain serves as a vital evolutionary mechanism, alerting the organism to potential harm or signaling the onset of illness. It plays a critical role in the body's defense system by helping preserve physiological integrity \cite{santiago_2022}. According to the biopsychosocial model, pain is not merely a physical sensation but the result of complex interactions among biological, psychological, and social factors \cite{cohen_vasem_hooten_2021}.
Pain has been described as a \textit{\textquotedblleft Silent Public Health Epidemic\textquotedblright} \cite{katzman_gallagher_2024}, emphasizing its widespread yet often underestimated impact. In nursing literature, it is also referred to as \textit{\textquotedblleft the fifth vital sign\textquotedblright} \cite{joel_lucille_1999}, reflecting the need for its routine and systematic assessment alongside other vital signs.
In the U.S., an estimated $50$ million people suffer daily from acute, chronic, or end-of-life pain, making it the leading reason for emergency room visits and medical consultations \cite{hhs_pain_2019}.
In addition, the widespread misuse of and addiction to pain medications have driven the opioid crisis in the United States, imposing substantial medical and social costs and causing numerous fatalities\cite{luo_li_2021}. In 2017 alone, opioid overdoses resulted in nearly $47,000$ fatalities \cite{wilson_kariisa_2020}.


The inherent subjectivity and complexity of pain assessment have been identified as significant challenges in both research and clinical practice. For example, pain management often relies on patients' subjective reports, making it difficult to administer medication with precision. This lack of objective evaluation contributes significantly to the overprescription and overuse of pain medications \cite{kong_chon_2024}.
Moreover, managing and assessing pain in patients with--or at risk of--medical instability presents significant clinical challenges, particularly when communication barriers are present \cite{puntilo_staannard_2022}. Research highlights that pain in critically ill adults remains frequently under-managed. A significant limitation is the lack of structured, comprehensive tools for assessing pain and supporting clinical decision-making in these contexts \cite{meehan_mcrae_1995}. Furthermore, cancer-related pain is highly prevalent, especially during advanced disease stages, where its occurrence exceeds $40\%$ \cite{bang_hak_2023}.

Pain assessment strategies span a broad spectrum. Self-reporting methods, including numerical rating scales and questionnaires, remain the gold standard for assessing patient experiences. In parallel, behavioral indicators---such as facial expressions, vocalizations, and body movements-are also used to infer pain, particularly in non-communicative patients \cite{rojas_brown_2023}. Physiological measures, such as electrocardiography (ECG), electromyography (EMG), and electrodermal activity (EDA), further enhance assessment by providing objective insights into the body's response to pain \cite{gkikas_tsiknakis_slr_2023}.
Electrodermal phenomena are primarily regulated by regions of the central nervous system involved in the classical sympathetic activation of sweat secretion. Nonetheless, a range of subcortical and cortical areas also contribute to the complex mechanisms governing EDA \cite{boucsein_1999}.
EDA captures changes in skin conductivity, which reflect sweat gland activity, and is widely regarded as a noninvasive indicator of sympathetic nervous system function. As such, it has gained prominence as a valuable physiomarker in both clinical and nonclinical contexts for evaluating SNS dynamics \cite{kong_chon_2024}.
Electrodermal activity has long been recognized as a reliable index for quantifying emotional responses \cite{traxel_1960}, stress-related arousal \cite{meijer_arts_2023}, and stress \cite{klimek_mannheim_2025}.
 In the domain of pain-related clinical applications, EDA is frequently utilized to evaluate the analgesic impact of treatments, therapies, or pharmacological interventions. For instance, \cite{perry_green_2018} reported a significant decrease in EDA integral values during therapy for low back pain, which corresponded with reduced pain intensity.

This study explores the application of electrodermal activity (EDA) as a standalone modality within an automatic pain assessment pipeline.
Although prior research has demonstrated the relevance of EDA in pain assessment and introduced various methods for processing, filtering, and modeling, the proposed pipeline integrates the most effective aspects of these approaches.
It introduces a parallel processing scheme that creates multiple representations of the original EDA signal. These representations are visualized as waveforms and fused into multi-representation diagrams, resulting in a multimodal-inspired framework based entirely on a single modality---EDA.

\section{Related Work}
Over the past $15$ years, interest in automatic pain assessment has steadily increased, with developments progressing from classical image and signal processing techniques to more advanced deep learning-based approaches \cite{gkikas_phd_thesis_2025}. The majority of existing methods are video-based, aiming to capture behavioral cues through facial expressions, body movements, or other visual indicators and employing a wide range of modeling strategies \cite{gkikas_tsiknakis_embc, bargshady_hirachan_2024,gkikas_tsiknakis_thermal_2024,huang_dong_2022}.
While video-based approaches dominate the field, a considerable number of studies have also focused on biosignal-based methods, although to a lesser extent. These works have investigated the utility of various physiological signals, such as electrocardiography (ECG) \cite{gkikas_chatzaki_2022, gkikas_chatzaki_2023}, electromyography (EMG) \cite{pavlidou_tsiknakis_2025,patil_patil_2024,thiam_bellmann_kestler_2019,werner_hamadi_niese_2014}, and brain activity through functional near-infrared spectroscopy (fNIRS) \cite{rojas_huang_2016, rojas_liao_2019,rojas_romero_2021,rojas_joseph_bargshady_2024,khan_sousani_2024,bargshady_aziz_2025}. 
In addition, multimodal approaches combining behavioral and physiological data have gained increasing attention in recent years, with several studies demonstrating the benefits of integrating multiple sources of information to improve performance \cite{farmani_giuseppi_2025,gkikas_rojas_painformer_2025,zhi_yu_2019,jiang_rosio_2024,jiang_li_he_2024,
gkikas_tachos_2024,farmani_bargshady_2025}.
For a comprehensive overview of methods employed in automatic pain recognition frameworks, refer to \cite{khan_umar_2025, gkikas_tsiknakis_slr_2023}.

Several studies have employed EDA signals for automatic pain assessment, spanning classical signal processing and feature engineering pipelines as well as deep-learning frameworks. 
For example, Aziz \textit{et al.} \cite{aziz_joseph_2025} used time-, spectral-, and cepstral-domain representations of EDA to extract features and trained several conventional classifiers. 
The authors in \cite{ji_zhao_li_2023} derived features from the frequency, time, and wavelet domains and employed ordinal-regression neural networks.
Interestingly, the study in \cite{sokolowska_mruzek_2025} analyzed both tonic and phasic EDA components, generating an extensive set of handcrafted features to compare electrical and pressure stimuli for pain recognition.
The authors in \cite{arenas_kong_2023} utilized phasic EDA components and developed a hybrid 1D model combining temporal convolutional networks with long short-term memory (LSTM) networks, achieving strong pain-detection performance. In \cite{li_luo_2025}, temporal convolution and cross-attention networks, together with multiscale time windows, delivered high pain-assessment accuracy. Similarly, Lu \textit{et al.} \cite{lu_ozek_2023} integrated squeeze-and-excitation residual networks with transformers and, through multiscale convolutional neural networks, exploited long, medium, and short temporal windows. Finally, \cite{phan_iyortsuun_2023} combined convolutional layers, BiLSTMs, and attention mechanisms, attaining state-of-the-art performance on benchmark datasets.

\section{Methodology}
This section outlines the signal pre-processing steps, the creation of multi-representation diagrams, and the architecture of the proposed model. It also provides details on the augmentation and regularization techniques used.

\subsection{Signal Pre-processing \& Feature Extraction}
\label{preprocessing}
Initially, the raw EDA signals $x(t)$ were low-pass filtered using a fourth-order Butterworth filter with a cutoff frequency of $5\,\text{Hz}$ and then linearly detrended to eliminate baseline drift \cite{rojas_hirachan_2023}.  
The resulting filtered signal $x_{\text{filt}}(t)$ was subsequently decomposed into tonic and phasic components using complementary $0.05\,\text{Hz}$ Butterworth low- and high-pass filters \cite{kim_pyo_2025}, defined as:
\begin{equation}
x_{\text{tonic}}(t) = \mathcal{L}_{0.05}\{x_{\text{filt}}(t)\}, \quad x_{\text{phasic}}(t) = x_{\text{filt}}(t) - x_{\text{tonic}}(t).
\end{equation}
The tonic component captures slow sympathetic tone, whereas the phasic component isolates rapid skin-conductance responses (SCRs).
Subsequently, residual drift in the phasic component was removed through linear detrending, resulting in a zero-baseline phasic waveform $x_{\text{pd}}(t)$, computed as:
\begin{equation}
  x_{\text{pd}}(t) = x_{\text{phasic}}(t) - \operatorname{mean}\bigl(x_{\text{phasic}}(t)\bigr).
\end{equation}
To calculate the time-varying sympathetic index (TVSymp), $x_{\text{pd}}(t)$ was downsampled and analysed using Variable-Frequency Complex Demodulation (VFCDM) \cite{quintero_florian_2016, wang_siu_2006}.  
Signal components within the $0.08$–$0.24\,\text{Hz}$ band---closely associated with sympathetic activation—were isolated, and their analytic amplitude was extracted using the Hilbert transform:
\begin{equation}
\text{TVSymp}(t) = \frac{\left|\mathcal{H} \left\{ \mathcal{B}_{0.08\text{--}0.24} \left( \text{VFCDM}\big(x_{\text{pd}}(t)\big) \right) \right\} \right|}{\sigma},
\end{equation}
where $\sigma$ denotes the standard deviation of the resulting envelope.  
The resulting waveform $\text{TVSymp}(t)$ serves as a dynamic index of sympathetic nervous system activity.
Lastly, a series of handcrafted features were extracted from the EDA signal to capture both temporal and spectral characteristics associated with sympathetic activity.  
Skin conductance responses (SCRs) were identified as local peaks in the phasic signal with amplitude $\ge40\%$ of the dynamic range and inter-peak interval $\ge0.5\,\text{s}$ \cite{benedek_kaernbach_2010, boucsein_2012}, yielding features such as SCR count, amplitude statistics, inter-peak intervals, and rise-time measures.
In addition, time-scale decomposition (TSD) features were computed over sliding windows of $1$, $2$, $4$, $5$, $8$, and $10\,\text{s}$ with $50\%$ overlap.  
For each window length $w$ and statistic $f$ (mean, standard deviation, slope, AUC, skewness, kurtosis), both the average and maximum values were calculated as:
\begin{align}
\overline{f}_w &= \frac{1}{K_w}\sum_{k=1}^{K_w} f_{k,w}, &
f^{\max}_w &= \max_{1 \le k \le K_w} f_{k,w},
\end{align}
where $\overline{f}_w$ the average of the statistic $f$ over all $K_w$ windows of length $w$, and $f^{\max}_w$ the maximum value of that statistic across those same windows.  
Spectral descriptors were also derived using Welch's method, including relative power in very-low ($0$–$0.045\,\text{Hz}$), low ($0.045$–$0.15\,\text{Hz}$), and high ($0.15$–$0.4\,\text{Hz}$) frequency bands, along with spectral entropy and centroid.  
Finally, distributional properties of the filtered signal were captured by extracting the $10$th, $25$th, $75$th, and $90$th percentiles.  
The handcrafted feature engineering process resulted in a feature vector of length $l = 98$.

\subsection{Multi-representation Diagrams}
\begin{figure*}
\begin{center}
\includegraphics[scale=0.32]{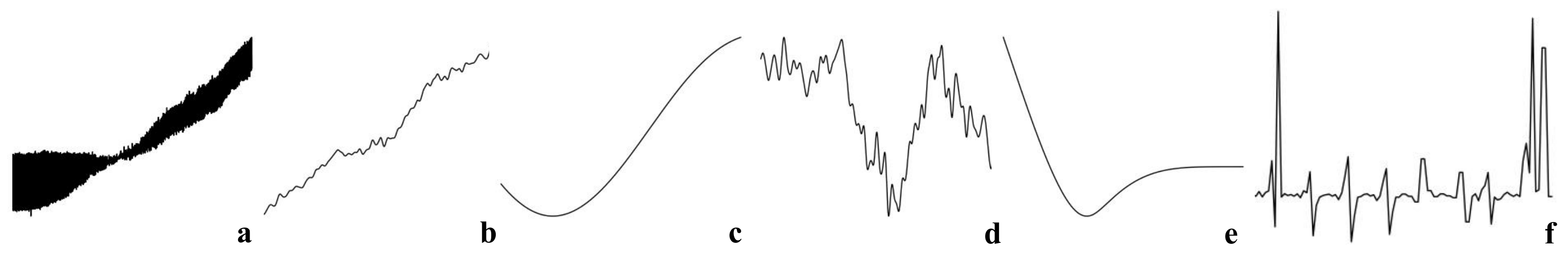}
\end{center}
\caption{Waveform diagram representations: (a) EDA-raw, (b) EDA-phasic, (c) EDA-tonic, (d) EDA-detrend, (e) EDA-TVSymp, and (f) EDA-handcrafted features.}
\label{waveforms}
\end{figure*}

\begin{figure}
\begin{center}
\includegraphics[scale=0.70]{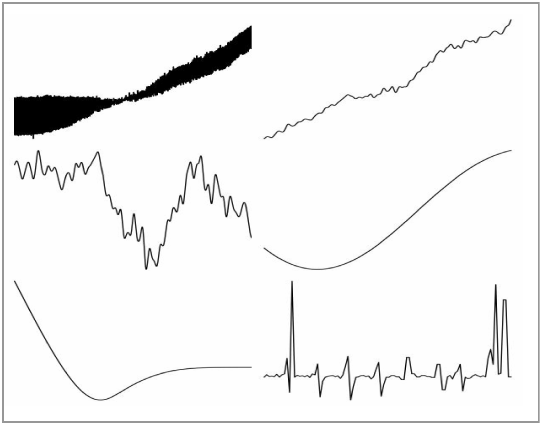}
\end{center}
\caption{Multi-representation diagram composed of six waveform representations.}
\label{multidiagram}
\end{figure}

The process described in \ref{preprocessing}, along with the original raw signal, resulted in six distinct representations of EDA: (1) \textit{EDA-raw}, (2) \textit{EDA-phasic}, (3) \textit{EDA-tonic}, (4) \textit{EDA-detrend}, (5) \textit{EDA-TVSymp}, and (6) \textit{handcrafted features}. These representations were subsequently visualized as waveform diagrams. A waveform diagram illustrates the temporal evolution of a signal, capturing its amplitude, frequency, and phase characteristics. It provides a simple visualization method, avoiding the need for transformations or complex computations such as those required for spectrograms, scalograms, or recurrence plots \cite{gkikas_tsiknakis_painvit_2024}.
Although the handcrafted feature vector is not a signal, it can still be visualized in the same way, as any 1D vector can be plotted in a 2D space. Each individual waveform diagram was then combined to form a multi-representation diagram with a resolution of $224\times224$ pixels.
Figure \ref{waveforms} presents the individual waveform representations, while Figure \ref{multidiagram} illustrates the combined multi-representation diagram.

\begin{figure*}
\begin{center}
\includegraphics[scale=0.375]{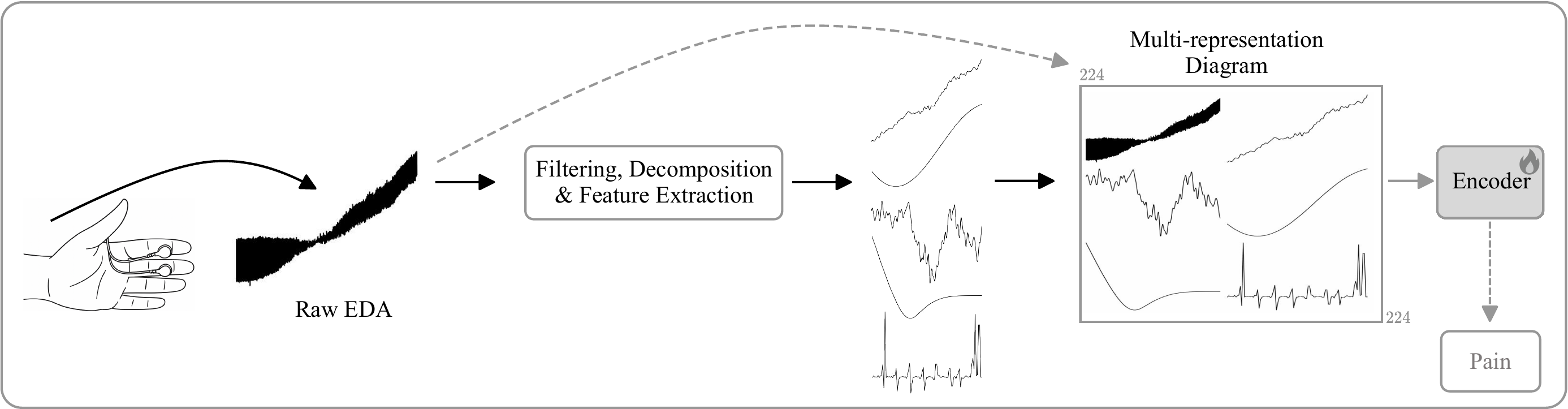}
\end{center}
\caption{Schematic overview of the proposed pipeline for pain assessment using EDA signals.}
\label{pipeline}
\end{figure*}

\subsection{Encoder}
The multi-representation diagrams are processed by a hierarchical Vision Transformer encoder, pretrained on approximately $11$ million images \cite{gkikas_rojas_painformer_2025}, which produces embeddings for the final pain classification.
Each diagram is divided into non-overlapping $16\times16$ patches, linearly projected to $d = 768$ tokens with positional encodings, and fused through alternating spectral-mixing and self-attention blocks.
The encoder architecture integrates two core components: spectral layers and self-attention layers.
In the spectral layer, input tokens $x \in \mathbb{R}^{h \times w \times d}$—where $h$, $w$, and $d$ denote the height, width, and channel dimension, respectively—are first transformed into the frequency domain using a two-dimensional Fast Fourier Transform (FFT), resulting in a complex tensor $X = \mathscr{F}[x] \in \mathbb{C}^{h \times w \times d}$.
A learnable complex filter $K$ is then applied element-wise ($\tilde{X} = K \odot X$) to modulate the frequency components. This modulated spectrum is transformed back to the spatial domain using an inverse FFT ($x \leftarrow \mathscr{F}^{-1}[\tilde{X}]$). The output is then passed through a channel-mixing MLP defined as $\Phi(x) = W_2 \cdot \text{GELU}(\text{DWConv}(W_1 \cdot x + b_1)) + b_2$, where $\text{DWConv}$ denotes a depthwise convolution. Layer normalization is applied both before and after the FFT and IFFT operations.
The self-attention layer follows the standard multi-head attention mechanism. For input $X$, attention is computed as:
\begin{equation}
\text{Att}(X) = \text{softmax}\left(\tfrac{XW_q (XW_k)^\mathsf{T}}{\sqrt{d}}\right) XW_v, 
\end{equation}
where $W_q$, $W_k$, and $W_v$ are the query, key, and value projection matrices, respectively. The output is then passed through a feedforward MLP: $\Phi(x) = W_2 \cdot \text{GELU}(W_1 \cdot x + b_1) + b_2$, with layer normalization applied before and after the attention mechanism.
The embedding produced by the encoder is finally passed through a fully connected layer to perform the classification task of pain recognition.
We note that the encoder has been fine-tuned in an end-to-end manner along with the entire pipeline.
Figure \ref{pipeline} presents an overview of the proposed pipeline.

\subsection{Augmentation Methods \& Regularization}
Several data augmentation techniques are applied during training. 
Every $224\times224$ multi-representation diagram undergoes a cascade of stochastic transformations.
The \textit{AugMix} method blends three randomly generated augmentation chains with the original image, shifting a mixture of contrast, colour, and geometric perturbations to the image. 
In addition, \textit{TrivialAugment} applies a single randomly chosen operation with a magnitude sampled uniformly. \textit{Centre cropping} is applied with a probability drawn from a given range, where the crop size is selected randomly and the image is resized back to its original dimensions. 
Conditional \textit{Gaussian blurring} applies noise by reducing high-frequency components.
Two \textit{Cutout} masks are applied—one placing a small number of blocks, the other covering the image with a higher number, both using blocks of equal size, $32\times32$.
Beyond data-level augmentation, two stochastic regularizers have been utilized. A \textit{Dropout} layer follows the encoder, with its keep probability gradually reduced linearly over training epochs:
\begin{equation}
p(t) \;=\; p_{\text{start}} + \frac{t}{T}\bigl(p_{\text{end}}-p_{\text{start}}\bigr),\qquad 0\le t\le T.
\end{equation}
The cross-entropy loss employs a \textit{Label-Smoothing} based on the same linear schedule, shifting from a soft target distribution toward one-hot labels.
Finally, a cosine learning-rate profile with \textit{Warmup} and \textit{Cooldown} schedules is also employed.
All stochastic choices---operation selection, magnitudes, cropping ratios, mask locations, and probability draws---are resampled independently for every image at every batch.
Throughout all experiments, the batch size is fixed to $32$ and the learning rate is set to $1\mathrm{e}{-4}$.

\section{Experimental Evaluation \& Results}
This study leverages the dataset released by the challenge organizers, which consists of electrodermal activity recordings from $65$ participants. Data collection took place at the Human-Machine Interface Laboratory, University of Canberra, Australia, and is divided into $41$ training, $12$ validation, and $12$ testing subjects. 
Pain stimulation was induced using transcutaneous electrical nerve stimulation (TENS) electrodes positioned on the inner forearm and the back of the right hand. 
Two pain levels were measured: pain threshold---the minimum stimulus intensity perceived as painful (low pain), and pain tolerance---the maximum intensity tolerated before becoming unbearable (high pain). 
The electrodermal activity sensors were attached to the left hand, with the two electrodes positioned on the proximal phalanges of the index and middle fingers.
The signals have a frequency of $100$ Hz and a duration of approximately $10$ seconds.
We refer to \cite{ai4pain_2025,rojas_hirachan_2023} for a detailed description of the recording protocol and to \cite{ai4pain_2024} for information regarding the previous edition of the challenge.
The experiments presented in this study are conducted on the validation subset of the dataset, evaluated under a multi-class classification framework with three levels: No Pain, Low Pain, and High Pain.
The validation results are reported in terms of macro-averaged accuracy, precision, and F1 score. The final results of the testing set are also reported. We note that all experiments followed a deterministic setup, eliminating the effect of random initializations; thus, any performance differences arose strictly from the chosen optimization settings, modalities, or other intentional changes rather than chance.

\subsection{Isolated Representations}
\begin{figure}
\begin{center}
\includegraphics[scale=0.13]{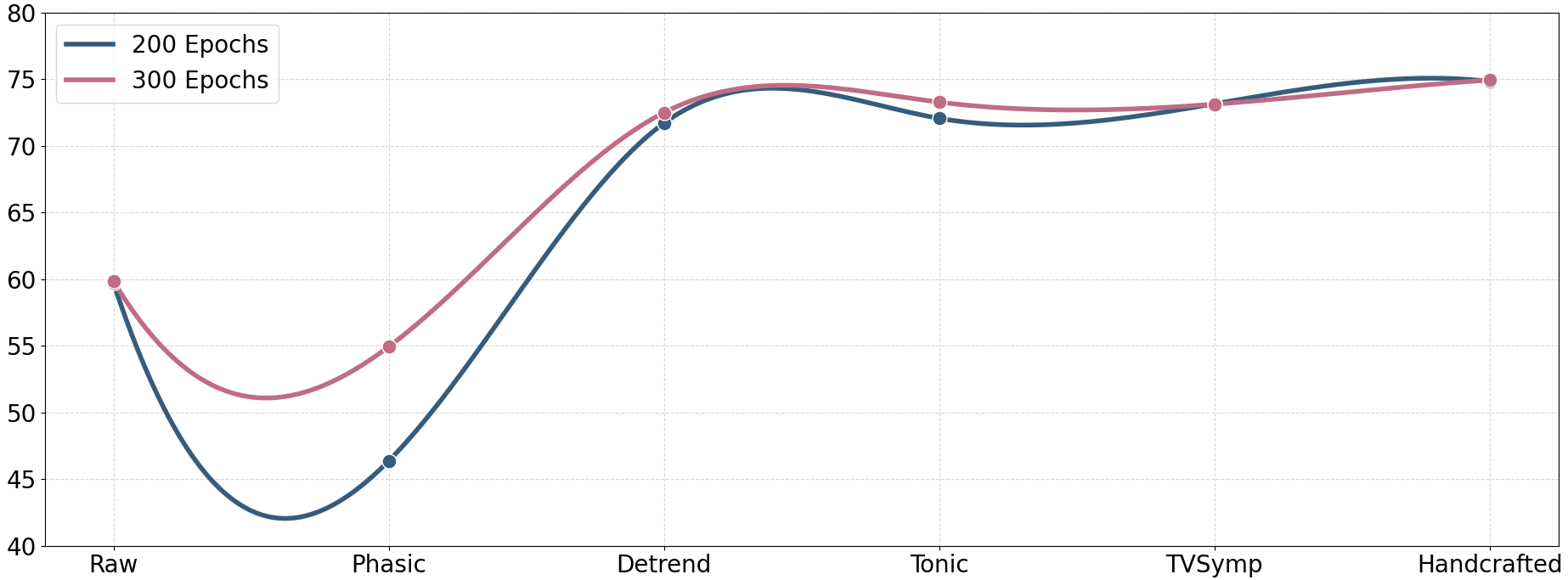}
\end{center}
\caption{Visual comparison of performance across different EDA representations and training epochs.}
\label{inputs_comprarison}
\end{figure}

In the context of representation analysis, each individual variant—namely, \textit{EDA-raw}, \textit{EDA-phasic}, \textit{EDA-tonic}, \textit{EDA-detrend}, \textit{EDA-TVSymp}, and \textit{handcrafted features}---was evaluated independently as an isolated diagram. Table \ref{table:representations} presents the corresponding results. The first set of experiments was conducted using a training duration of $200$ epochs, with the first $30$ epochs utilized for warmup. We observe an accuracy of $59.73\%$ for the raw EDA signals, while the phasic representation achieved only $46.36\%$, where training collapse was also observed. The detrended signals performed significantly better, achieving $71.71\%$ accuracy, followed closely by the tonic representation at $72.07\%$. The highest overall performance was achieved by the final two representations: TVSymp with $73.16\%$ accuracy, $75.19\%$ precision, and $73.90\%$ F1 score, while the handcrafted features reached $74.84\%$, $73.46\%$, and $73.59\%$ respectively.
The second set of experiments extended the training duration to 300 epochs, with a $50$-epoch warmup, based on the observation that the representations could benefit from longer and more stable training. While performance improved overall, the increase was not uniform across all inputs. Raw EDA saw a marginal improvement of $0.13\%$ in accuracy. The phasic signals, however, showed the most notable gain, rising to $54.93\%$ accuracy and avoiding the training instability observed earlier. Detrended, tonic, and handcrafted representations saw respective increases of $0.79\%$, $1.21\%$, and $0.11\%$, with the handcrafted features again achieving the highest accuracy at $74.95\%$. TVSymp showed a modest decrease across all three metrics.
Figure \ref{inputs_comprarison} provides a visual summary of the performance across all representations under both the 200- and $300$-epoch configurations.

\begin{table}
\caption{Comparison of performances for different EDA representations.}
\label{table:representations}
\begin{center}
\begin{threeparttable}
\begin{tabular}{ P{1.0cm} P{1.2cm} P{1.5cm} P{1.05cm} P{0.91cm} P{0.61cm}}
\toprule
\multirow{2}[2]{*}{\shortstack{Epochs}}
&\multirow{2}[2]{*}{\shortstack{Input}}
&\multicolumn{1}{c}{Schedule} 
&\multicolumn{3}{c}{Task--MC}\\ 
\cmidrule(lr){3-3}\cmidrule(lr){4-6} 
& &Warmup &Accuracy &Precision &F1\\
\midrule
\midrule
200 &Raw         &30 &59.73             &64.17             &60.69\\
200 &Phasic      &30 &46.36             &41.90             &44.02\\
200 &Detrend     &30 &71.71             &72.87             &72.07\\
200 &Tonic       &30 &72.07             &\underline{74.01} &72.97\\
200 &TVSymp      &30 &\underline{73.16} &\textbf{75.19}    &\textbf{73.90}\\
200 &Handcrafted &30 &\textbf{74.84}    &73.46             &\underline{73.59}\\\midrule
300 &Raw         &50 &59.86             &68.97             &63.40\\
300 &Phasic      &50 &54.93             &57.40             &56.04\\
300 &Detrend     &50 &72.50             &73.13             &72.80\\
300 &Tonic       &50 &\underline{73.28} &\textbf{75.01}    &73.80\\
300 &TVSymp      &50 &73.13             &\underline{74.68} &\underline{73.87}\\
300 &Handcrafted &50 &\textbf{74.95}    &73.53             &\textbf{74.16}\\

\bottomrule 
\end{tabular}
\begin{tablenotes}[para,flushleft] 
\scriptsize                   
\item \textbf{MC}: multiclass classification (No, Low, High Pain)
\end{tablenotes}
\end{threeparttable}
\end{center}
\end{table}

\subsection{Fusion of Representations}
\begin{figure}
\begin{center}
\includegraphics[scale=0.125]{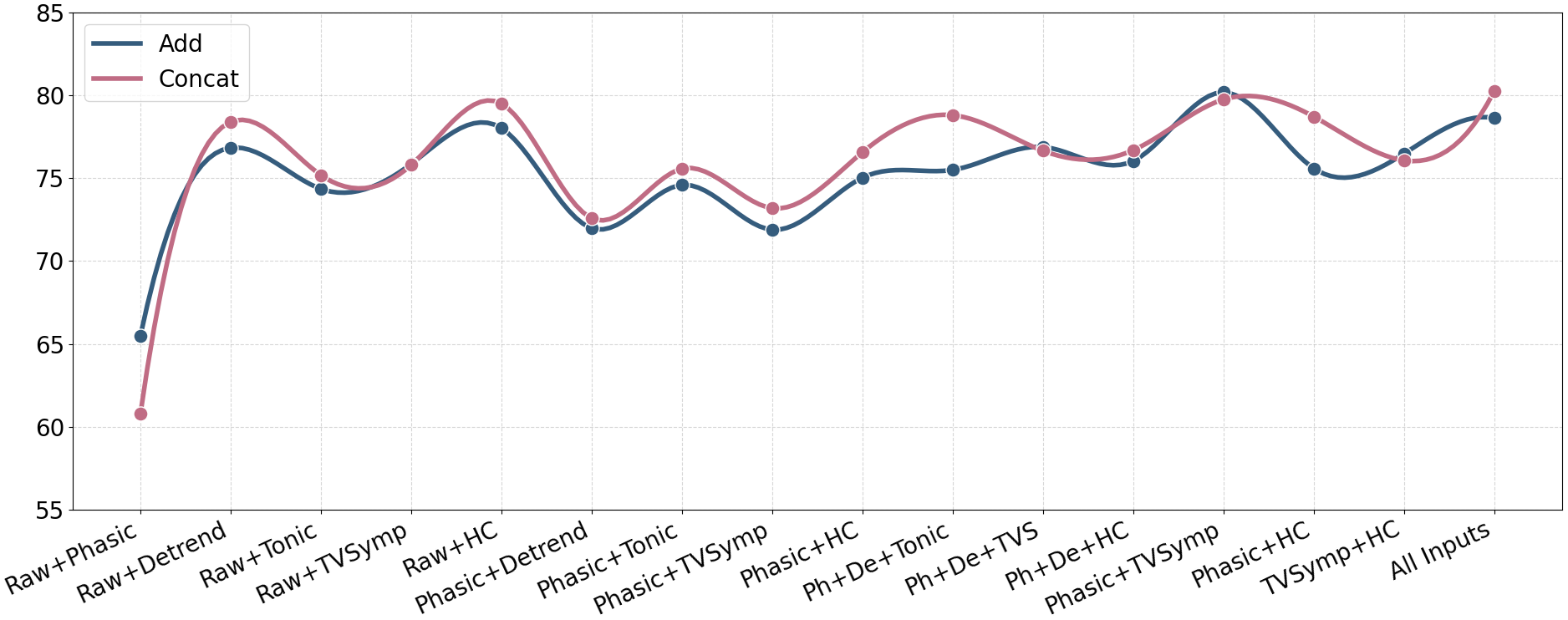}
\end{center}
\caption{Visual comparison of performance across different combinations of EDA representations using the two fusion methods: addition and concatenation.}
\label{fusion}
\end{figure}

The first category of fusion applied to the representations is late fusion, a classic technique. In this scheme, the encoder extracts embeddings from each individual representation, and simple operations such as addition or concatenation are employed to combine them before classification. The results are summarized in Table \ref{table:fusion}.
Initially, combinations involving the raw signal and individual representations were evaluated using addition. Fusion with the phasic representation yielded an accuracy of $65.50\%$, with detrended signals achieving $76.83\%$, and tonic resulted in $74.35\%$. Combining raw with TVSymp reached $75.86\%$, while raw with handcrafted features achieved $78.01\%$. When using concatenation for the same representation pairs, performance was generally lower on average; however, peak performance was higher. For instance, raw and detrended dropped to $71.98\%$, while raw and tonic increased to $78.39\%$. The combination of raw and handcrafted data via concatenation produced the highest overall performance in this group, with $79.49\%$ accuracy, $80.43\%$ precision, and $79.61\%$ F1 score.
The second group of combinations focused on the phasic representation paired with others. Using addition, phasic with detrend reached $71.98\%$, with tonic $74.59\%$, and with handcrafted $75.03\%$. In this group, concatenation consistently outperformed addition, with observed increases of $0.60\%$, $0.98\%$, $1.30\%$, and $1.54\%$ for the phasic–detrend, phasic–tonic, phasic–TVSymp, and phasic–handcrafted pairs, respectively. The phasic–handcrafted pair with concatenation yielded the best result in this group, with $76.57\%$ accuracy, $76.41\%$ precision, and $76.38\%$ F1 score.
Further combinations involved three-input setups. Using addition, phasic–detrend–tonic achieved $75.50\%$, phasic–detrend–TVSymp $76.88\%$, and phasic–detrend–handcrafted $76.04\%$. The concatenation variant of phasic–detrend–tonic reached $78.80\%$, which was the highest among all three-input combinations.
Other evaluated combinations included tonic–TVSymp and phasic–handcrafted. Using addition, these achieved $80.18\%$ and $75.59\%$ accuracy, respectively---the highest for their category. With concatenation, tonic–TVSymp experienced a slight drop, while phasic–handcrafted improved to $78.71\%$.
In contrast, the TVSymp and handcrafted pair exhibited relatively lower performance, reaching a maximum of $76.48\%$ via addition.
Finally, combining all six available inputs yielded the best overall results. Concatenation yielded $80.27\%$ accuracy, $81.37\%$ precision, and $80.43\%$ F1 score. The addition-based approach, while slightly behind in overall accuracy and F1, achieved the highest precision at $82.56\%$.
Figure \ref{fusion} provides a visual comparison of the performance across the different input combinations and fusion methods discussed above.

\begin{table}
\caption{Comparison of performance across different EDA representation combinations and fusion methods.}
\label{table:fusion}
\begin{center}
\begin{threeparttable}
\begin{tabular}{ P{0.8cm} P{2.1cm} P{0.8cm} P{1.05cm} P{0.91cm} P{0.61cm}}
\toprule
\multirow{2}[2]{*}{\shortstack{Epochs}}
&\multirow{2}[2]{*}{\shortstack{Input}}
&\multirow{2}[2]{*}{\shortstack{Fusion}}
&\multicolumn{3}{c}{Task--MC}\\ 
\cmidrule(lr){4-6} 
& & &Accuracy &Precision &F1\\
\midrule
\midrule
300 &Raw, Phasic     &add    &65.50             &73.42 &68.80\\
300 &Raw, Detrend    &add    &76.83             &78.70 &77.70\\
300 &Raw, Tonic      &add    &74.35             &76.38 &75.30\\
300 &Raw, TVSymp     &add    &75.86             &76.11 &75.71\\
300 &Raw, HC         &add    &78.01             &79.09 &78.44\\\hdashline
300 &Raw, Phasic     &concat &60.80             &61.07 &60.90\\
300 &Raw, Detrend    &concat &\underline{78.39} &79.77 &\underline{79.00}\\
300 &Raw, Tonic      &concat &75.18             &\underline{80.38} &77.56\\
300 &Raw, TVSymp     &concat &75.80             &78.83 &75.84\\
300 &Raw, HC         &concat &\textbf{79.49}    &\textbf{80.43} &\textbf{79.61}\\\midrule

300 &Phasic, Detrend &add    &71.98             &71.64             &71.50\\
300 &Phasic, Tonic   &add    &74.59             &73.75             &74.13\\
300 &Phasic, TVSymp  &add    &71.89             &72.64             &71.43\\
300 &Phasic, HC      &add    &75.03             &75.64             &75.33\\\hdashline
300 &Phasic, Detrend &concat &72.58             &74.20             &73.36\\
300 &Phasic, Tonic   &concat &\underline{75.57} &\underline{75.65} &\underline{75.58}\\
300 &Phasic, TVSymp  &concat &73.19             &74.58             &73.76\\
300 &Phasic, HC      &concat &\textbf{76.57}    &\textbf{76.41}    &\textbf{76.38}\\\midrule

\multirow{2}{*}{300} &
\multirow{2}{*}{\shortstack{Phasic, Detrend,\\Tonic}} &
\multirow{2}{*}{add} &
\multirow{2}{*}{\centering 75.50} &
\multirow{2}{*}{\centering 76.62} &
\multirow{2}{*}{\centering 76.03} \\\\

\multirow{2}{*}{300} &
\multirow{2}{*}{\shortstack{Phasic, Detrend,\\TVSymp}} &
\multirow{2}{*}{add} &
\multirow{2}{*}{\centering \underline{76.88}} &
\multirow{2}{*}{\centering 77.64} &
\multirow{2}{*}{\centering \underline{77.21}} \\\\

\multirow{2}{*}{300} &
\multirow{2}{*}{\shortstack{Phasic, Detrend,\\HC}} &
\multirow{2}{*}{add} &
\multirow{2}{*}{\centering 76.04} &
\multirow{2}{*}{\centering 76.58} &
\multirow{2}{*}{\centering 76.29} \\\\\hdashline

\multirow{2}{*}{300} &
\multirow{2}{*}{\shortstack{Phasic, Detrend,\\Tonic}} &
\multirow{2}{*}{concat} &
\multirow{2}{*}{\centering \textbf{78.80}} &
\multirow{2}{*}{\centering \textbf{78.39}} &
\multirow{2}{*}{\centering \textbf{78.55}} \\\\

\multirow{2}{*}{300} &
\multirow{2}{*}{\shortstack{Phasic, Detrend,\\TVSymp}} &
\multirow{2}{*}{concat} &
\multirow{2}{*}{\centering 76.65} &
\multirow{2}{*}{\centering \underline{77.66}} &
\multirow{2}{*}{\centering 76.32} \\\\

\multirow{2}{*}{300} &
\multirow{2}{*}{\shortstack{Phasic, Detrend,\\HC}} &
\multirow{2}{*}{concat} &
\multirow{2}{*}{\centering 76.69} &
\multirow{2}{*}{\centering 76.86} &
\multirow{2}{*}{\centering 74.96} \\\\\midrule

300 &Tonic, TVSymp &add    &\textbf{80.18}    &\textbf{82.50}    &\textbf{80.62}\\
300 &Tonic, HC     &add    &75.59             &77.31             &76.11\\\hdashline
300 &Tonic, TVSymp &concat &\underline{79.75} &\underline{79.41} &\underline{79.38}\\
300 &Tonic, HC     &concat &78.71             &78.76             &78.69\\\midrule

300 &TVSymp, HC  &add    &\textbf{76.48} &76.97          &\textbf{76.40}\\\hdashline
300 &TVSymp, HC  &concat &76.07          &\textbf{77.29} &76.23\\\midrule

\multirow{3}{*}{300} &
\multirow{3}{*}{\shortstack{Raw, Phasic, \\Detrend, Tonic, \\TVSymp, HC}} &
\multirow{3}{*}{add} &
\multirow{3}{*}{\centering 78.62} &
\multirow{3}{*}{\centering \textbf{82.56}} &
\multirow{3}{*}{\centering 78.65} \\\\\\\hdashline

\rowcolor{mygray}
300 & \makecell{Raw, Phasic,\\Detrend, Tonic\\TVSymp, HC} & concat
    & \textbf{80.27} & 81.37 & \textbf{80.43} \\

\bottomrule 
\end{tabular}
\begin{tablenotes}[para,flushleft] 
\scriptsize                   
\item \textbf{HC}: handcrafted feature diagram
\end{tablenotes}
\end{threeparttable}
\end{center}
\end{table}

\subsection{Multi-representation Diagrams}
\begin{figure}
\begin{center}
\includegraphics[scale=0.1235]{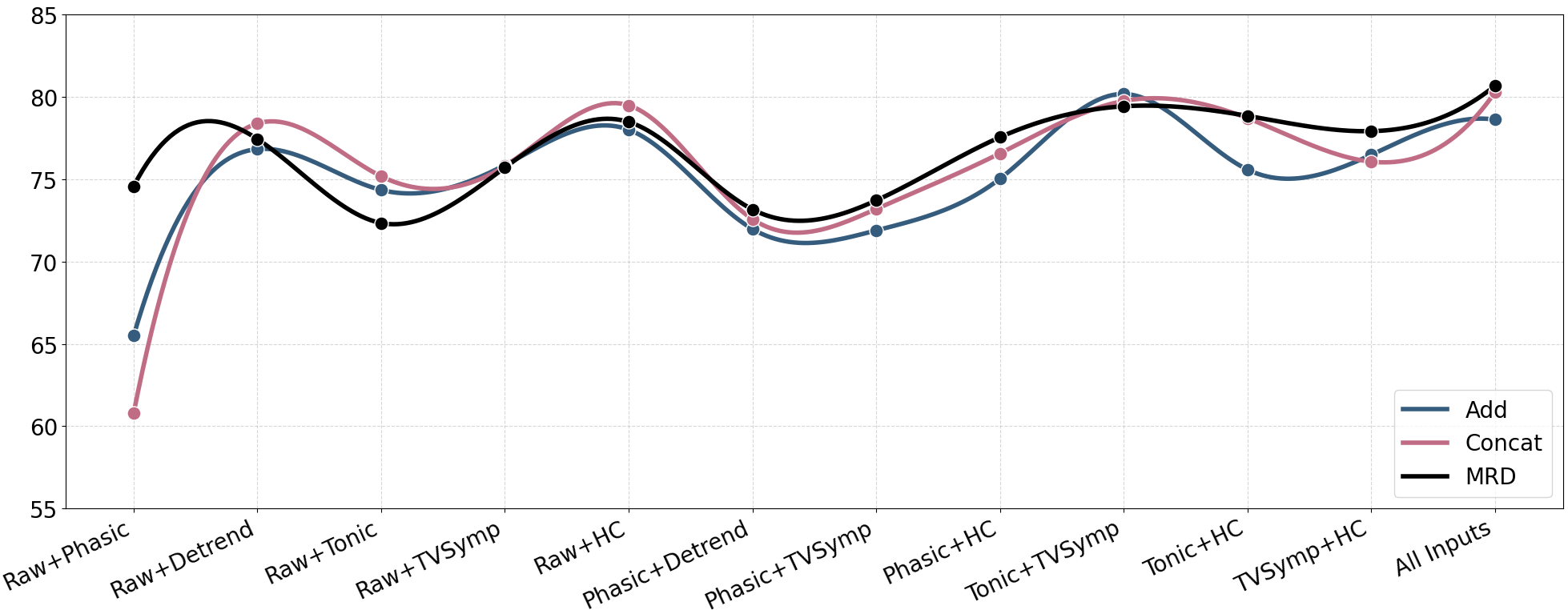}
\end{center}
\caption{Visual comparison of performance across different combinations of EDA representations using the three fusion methods: addition, concatenation, and multi-representation diagrams.}
\label{final_comparison}
\end{figure}

The proposed multi-representation diagrams were evaluated by combining different waveform representations, ranging from pairs to the inclusion of all six available types. Note that when only two representations are used, they are visualized vertically within the diagram, one placed above the other. The corresponding results are presented in Table \ref{table:diagrams}.
Visualizing the raw and phasic signals within a single diagram resulted in an accuracy of $74.58\%$, a substantial improvement over the classic fusion approaches, exceeding addition and concatenation by more than 9 and 13 percentage points, respectively. The raw--detrend combination reached $77.44\%$ accuracy, outperforming the addition method, but slightly underperforming compared to concatenation. For the raw--tonic pair, an accuracy of $72.34\%$ was recorded, approximately $2\%$ lower than both classic fusion baselines. The raw-TVSymp fusion achieved $75.71\%$, closely matching the corresponding values from the standard fusion methods. Notably, combining raw and handcrafted representations yielded $78.78\%$ accuracy, higher than addition, but slightly below concatenation.
In the case of phasic and detrend, the combined diagram achieved $73.17\%$ accuracy, improving upon the classic methods by approximately $1\%$. For the phasic--TVSymp pair, $73.71\%$ was recorded, again roughly $1\%$ higher than the baseline methods. However, the phasic--tonic combination experienced training instability, resulting in a collapse of learning and, consequently, no meaningful performance. When using the handcrafted representation with phasic, accuracy rose to $77.55\%$, surpassing both classic methods by approximately $1.5\%$.
Additional combinations of detrend with tonic, TVSymp, and handcrafted representations yielded accuracies of $77.30\%$, $76.71\%$, and $76.85\%$, respectively. Although no direct classic fusion counterparts were available for these specific combinations, performance was consistently high. The tonic--TVSymp combination achieved $79.43\%$ accuracy, slightly below ($0.75\%$) the best result from concatenation. In contrast, tonic--handcrafted reached $78.85\%$, marginally outperforming its concatenation counterpart by $0.14\%$. The TVSymp--handcrafted pairing reported $77.92\%$ accuracy, which is $1.44\%$ higher than the best corresponding addition-based result.
Finally, combining all six waveform representations within a single diagram yielded the highest overall performance: $80.67\%$ accuracy, $81.74\%$ precision, and $80.89\%$ F1 score. These results exceed those obtained through the best classic fusion method (concatenation), which achieved $80.27\%$ accuracy.
Figure \ref{final_comparison} presents a comparative overview of performance across the three fusion strategies---addition, concatenation, and the proposed multi-representation diagrams. Note that only the common representation combinations across all methods are included.

\begin{table}
\caption{Comparison of performance across different EDA representation combinations and the proposed diagrams.}
\label{table:diagrams}
\begin{center}
\begin{threeparttable}
\begin{tabular}{ P{0.6cm} P{2.4cm} P{0.8cm} P{1.04cm} P{0.90cm} P{0.60cm}}
\toprule
\multirow{2}[2]{*}{\shortstack{Epochs}}
&\multirow{2}[2]{*}{\shortstack{Input}}
&\multirow{2}[2]{*}{\shortstack{Fusion}}
&\multicolumn{3}{c}{Task--MC}\\ 
\cmidrule(lr){4-6} 
& & &Accuracy &Precision &F1\\
\midrule
\midrule
300 &Raw, Phasic     &MRD &74.58             &77.08             &75.53\\
300 &Raw, Detrend    &MRD &\underline{77.44} &\underline{78.26} &\underline{77.82}\\
300 &Raw, Tonic      &MRD &72.34             &71.35             &71.79\\
300 &Raw, TVSymp     &MRD &75.71             &75.77             &75.69\\
300 &Raw, HC         &MRD &\textbf{78.48}    &\textbf{79.54}    &\textbf{78.82}\\\midrule

300 &Phasic, Detrend &MRD &73.17             &70.79             &71.85\\
300 &Phasic, Tonic   &MRD &56.78             &59.70             &57.30\\
300 &Phasic, TVSymp  &MRD &\underline{73.71} &\underline{76.25} &\underline{74.26}\\
300 &Phasic, HC      &MRD &\textbf{77.55}    &\textbf{77.37}    &\textbf{77.06}\\\midrule

300 &Detrend, Tonic  &MRD &\textbf{77.30}    &76.00             &76.58\\
300 &Detrend, TVSymp &MRD &76.71             &\underline{76.60} &\underline{76.59}\\
300 &Detrend, HC     &MRD &\underline{76.85} &\textbf{78.08}    &\textbf{77.41}\\\midrule

300 &Tonic, TVSymp  &MRD  &\textbf{79.43} &\textbf{81.13} &\textbf{80.22}\\
300 &Tonic, HC      &MRD  &78.85          &78.80          &78.84\\\midrule

300 &TVSymp, HC     &MRD  &77.92             &78.36             &78.02\\\midrule

\rowcolor{mygray}
300 & \makecell{Raw, Phasic,\\Detrend, Tonic\\TVSymp, HC} & MRD
    & \textbf{80.67} & \textbf{81.74} & \textbf{80.89} \\

\bottomrule 
\end{tabular}
\begin{tablenotes}[para,flushleft] 
\scriptsize                   
\item \textbf{MRD}: multi-representation diagram
\end{tablenotes}
\end{threeparttable}
\end{center}
\end{table}

\subsection{Optimization for the Multi-representation Diagrams}
\begin{figure*}
\begin{center}
\includegraphics[scale=0.35]{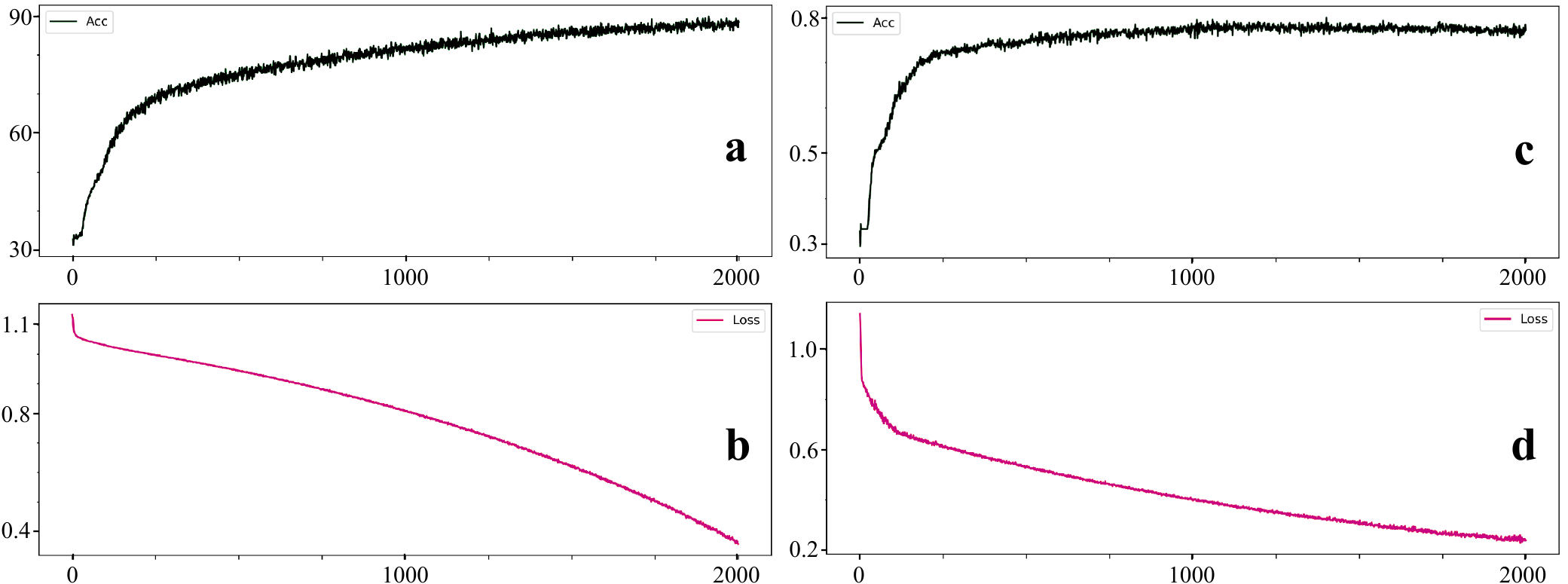}
\end{center}
\caption{Visualization of the performance of the final version of the proposed method: (a) training accuracy, (b) validation accuracy, (c) training loss, and (d) validation loss.
}
\label{final_performances}
\end{figure*}

The final series of experiments focused on optimizing the previously reported results using the multi-representation diagrams. Prior findings had already demonstrated strong performance, particularly when all six EDA waveform representations were combined. These experiments aimed to refine the training pipeline by exploring training schedules and regularization strategies, targeting not only peak performance but also improved stability. The results are presented in Table \ref{table:final}.
Initially, increasing the \textit{Label Smoothing} probability to $30\%$ and $70\%$ yielded accuracies of $81.63\%$ and $80.93\%$, respectively---the former showing an improvement of over $1\%$ compared to the baseline. Applying a linear decay to the label smoothing from $70\%$ to $10\%$ resulted in an accuracy of $80.65\%$. Keeping the smoothing fixed at $70\%$ while applying a linear decrease to \textit{Dropout} from $90\%$ to $10\%$ led to $81.31\%$. 
The best performance was achieved when both \textit{Label Smoothing} and \textit{Dropout} were scheduled to decrease linearly, resulting in an accuracy of $82.03\%$. These results suggest that dynamic regularization strategies improve learning effectiveness.
Subsequent experiments explored further combinations of regularization and scheduling mechanisms, such as \textit{Warmup} and learning rate adjustments. However, these did not result in significant performance gains.
Finally, it was observed that many of the earlier configurations exhibited performance spikes and fluctuations during training. To address this, a stable training schedule was implemented, where training was extended to $2000$ epochs using a very low learning rate of $1\times10^{-6}$, along with strong regularization schedules---label smoothing decreasing from $70\%$ to $10\%$ and dropout from $90\%$ to $50\%$. This configuration yielded an accuracy of $79.92\%$, with precision and F1 scores of $80.69\%$ and $80.23\%$, respectively. While this was not the highest in terms of peak performance, it resulted in the smoothest and most stable training process. The training and validation curves are shown in Figure \ref{final_performances}.

\begin{table}
\caption{Comparison of various optimization strategies applied to the multi-representation diagram using all six available waveform representations.}
\label{table:final}
\begin{center}
\begin{threeparttable}
\begin{tabular}{ P{1.0cm} P{0.90cm}  P{0.90cm} P{0.95cm}  P{0.95cm} P{1.8cm}  }
\toprule
\multirow{2}[2]{*}{\shortstack{Epochs}}
&\multicolumn{2}{c}{Schedule} 
&\multicolumn{2}{c}{Regularization} 
&\multicolumn{1}{c}{Task--MC}\\ 
\cmidrule(lr){2-3}\cmidrule(lr){4-5}\cmidrule(lr){6-6}  
&Warmup &LR &LS &DO  &Metrics\\
\midrule
\midrule
300 &50 &1e-4 &10-10 &50-50 &80.67\scriptsize{\textbar 81.74 \textbar 80.89}\\
300 &50 &1e-4 &30-30 &50-50 &\underline{81.63}\scriptsize{\textbar 82.56 \textbar 81.41}\\
300 &50 &1e-4 &70-70 &50-50 &80.93\scriptsize{\textbar 81.00 \textbar 80.92}\\\hdashline

300 &50 &1e-4 &70-10 &50-50  &80.65\scriptsize{\textbar 82.08 \textbar 80.42}\\
300 &50 &1e-4 &70-70  &90-10 &81.31\scriptsize{\textbar 81.88 \textbar 81.00}\\
300 &50 &1e-4 &70-10 &70-10  &\textbf{82.03}\scriptsize{\textbar 82.14 \textbar 81.90}\\\hdashline

300 &50 &1e-5 &70-10 &50-50 &79.95\scriptsize{\textbar 80.48 \textbar 80.15}\\
300 &50 &1e-5 &70-70 &90-10 &79.05\scriptsize{\textbar 80.39 \textbar 79.58}\\
300 &50 &1e-5 &70-10 &70-10 &78.04\scriptsize{\textbar 78.81 \textbar 78.25}\\\hdashline

300 &150 &1e-5 &70-10 &50-50 &79.31\scriptsize{\textbar 80.25 \textbar 79.70}\\
300 &150 &1e-5 &70-70 &90-10 &79.08\scriptsize{\textbar 81.58 \textbar 80.25}\\
300 &150 &1e-5 &70-10 &70-10 &79.20\scriptsize{\textbar 79.97 \textbar 79.35}\\\hdashline

300 &150 &1e-4 &70-10 &50-50 &79.72\scriptsize{\textbar 80.06 \textbar 79.56}\\
300 &150 &1e-4 &70-70 &90-10 &80.21\scriptsize{\textbar 83.43 \textbar 79.70}\\
300 &150 &1e-4 &70-10 &70-10 &80.01\scriptsize{\textbar 80.88 \textbar 80.39}\\\hdashline

\rowcolor{mygray} 2000 &10 &1e-6 &70-10 &90-50 &79.92\scriptsize{\textbar 80.69 \textbar 80.23}\\

\bottomrule 
\end{tabular}
\begin{tablenotes}[para,flushleft] 
\scriptsize                   
\item Values in the format \textbf{$x$--$y$} indicate that the corresponding parameters were linearly scheduled during training from $x\%$ to $y\%$.
\textbf{LR}: learning rate \textbf{LS}: label smoothing (in \%) \textbf{DO}:dropout rate (in \%). 
\end{tablenotes}
\end{threeparttable}
\end{center}
\end{table}

\section{Comparison with Existing Methods}
In this section, the proposed approach is compared with previous studies using the testing set of the \textit{AI4PAIN} dataset. Some of these studies were conducted as part of the \textit{First Multimodal Sensing Grand Challenge}. In contrast, others, including the present work, utilized data from the \textit{Second Multimodal Sensing Grand Challenge}. The key distinction between the two challenges lies in the set of available modalities.
Studies employing facial video or fNIRS have reported strong results, with accuracies of $49.00\%$ by \cite{prajod_schiller_2024} and $55.00\%$ by \cite{nguyen_yang_2024}, respectively. Combining these two modalities also yielded competitive results, although not substantially better than using each modality alone. For instance, \cite{vianto_2025} reported $51.33\%$, and \cite{gkikas_rojas_painformer_2025} achieved $55.69\%$ using fused video and fNIRS data.
Regarding the physiological modalities available in the \textit{Second Grand Challenge}, some approaches achieved particularly high performance, while others showed more limited results. In \cite{gkikas_kyprakis_multimodal_2025}, the authors reported $54.89\%$ accuracy using a combination of EDA, BVP, respiration, and blood oxygen saturation (SpO$_2$). In contrast, \cite{gkikas_kyprakis_resp_2025} achieved $42.17\%$ using only the respiration signal.
The proposed method, based solely on EDA, achieved an accuracy of $55.17\%$---among the highest reported to date. This outcome reinforces prior findings, as EDA is widely recognized as one of the most effective modalities for pain assessment and stress-related applications, as outlined in the introduction.
Table \ref{table:ai4pain_test} presents the corresponding results and comparisons across the studies.

\begin{table}
\caption{Comparison of studies on the testing set of the \textit{AI4Pain} dataset.}
\label{table:ai4pain_test}
\begin{center}
\begin{threeparttable}
\begin{tabular}{P{0.7cm} P{2.0cm} P{3.0cm} P{1.0cm}}
\toprule
Study & Modality & ML & Acc (\%) \\
\midrule
\midrule
\cite{khan_aziz_2025}$^\dagger$                &fNIRS        &ENS             &53.66\\ \hdashline
\cite{nguyen_yang_2024}$^\dagger$              &fNIRS        &Transformer     &55.00\\ \hdashline
\cite{prajod_schiller_2024}$^\dagger$          &Video        &2D CNN          &49.00\\ \hdashline
\cite{gkikas_tsiknakis_painvit_2024}$^\dagger$ &Video, fNIRS &Transformer     &46.67\\ \hdashline
\cite{vianto_2025}$^\dagger$                   &Video, fNIRS &CNN-Transformer &51.33 \\\hdashline
\cite{gkikas_rojas_painformer_2025}$^\dagger$  &Video, fNIRS &Transformer     &55.69 \\\midrule

\cite{gkikas_kyprakis_multimodal_2025}$^\ddagger$    &EDA, BVP, Resp, SpO$_2$ &MoE     &54.89\\ \hdashline
\cite{gkikas_kyprakis_resp_2025}$^\ddagger$    &Respiration             &Transformer &42.24 \\\hdashline
Our$^\ddagger$        &EDA                     &Transformer &55.17\\

\bottomrule 
\end{tabular}
\begin{tablenotes}
\scriptsize
\item \textbf{ENS}: Ensemble Classifier  \textbf{SpO$_2$}: Peripheral Oxygen Saturation \textbf{MoE}: Mixture of Experts 
$\pmb{\dagger}$: AI4PAIN-First Multimodal Sensing Grand Challenge $\pmb{\ddagger}$: AI4PAIN-Second Multimodal Sensing Grand Challenge
\end{tablenotes}
\end{threeparttable}
\end{center}
\end{table}

\section{Discussion \& Conclusion}
This study presents our contribution to the \textit{Second Multimodal Sensing Grand Challenge for Next-Generation Pain Assessment (AI4PAIN)}, where electrodermal activity (EDA) signals were the chosen modality.
A novel approach was proposed for combining different EDA signal representations into a single multi-representation diagram. This method was evaluated against classical late fusion strategies, specifically feature addition and concatenation. The results demonstrated that the proposed method achieved comparable and, in some cases, superior performance, indicating its potential as a robust alternative in scenarios requiring the integration of multiple signal representations.
Additionally, the effectiveness of visualizing a smaller number of representations—either single waveforms or pairs—was also examined. These configurations showed competitive performance compared to their counterparts using traditional feature fusion or representation techniques.
Overall, the findings highlight that transforming signal waveforms into image-based visualizations can be a highly effective strategy. This challenges the conventional reliance on 1D signal analysis and opens the possibility of leveraging powerful 2D vision models, which are known for their ability to extract high-quality local features.
Beyond waveform visualizations, this work also explored the transformation of EDA-derived feature vectors into visual representations. These diagrams demonstrated strong performance both independently and in combination with other inputs, suggesting the method's applicability in feature-based approaches as well.
One potential limitation of the proposed method lies in its scalability. While experiments covered up to six representations, using significantly more (\textit{e.g.}, ten, twelve, or more) may lead to excessive compression within the diagram, which inevitably degrades performance.
The challenge results confirmed the competitiveness of the proposed approach relative to other studies. This reflects the synergy between the high-performing EDA modality and the effectiveness of the multi-representation diagram approach.
In conclusion, multi-representation diagrams offer a promising alternative for integrating diverse representations—whether signal-based or feature-based—particularly in settings where the total number remains moderate.

\section*{Safe and Responsible Innovation Statement}
This work relied on the \textit{AI4PAIN} dataset \cite{rojas_hirachan_2023,ai4pain_2024,ai4pain_2025}, made available by the challenge organizers, to assess automatic pain recognition methods. All participants confirmed the absence of neurological or psychiatric conditions, unstable health issues, chronic pain, or regular medication use during the session. Before the experiment, participants were thoroughly informed of the procedures, and written consent was obtained. The original study's human-subject protocol received ethical clearance from the University of Canberra's Human Ethics Committee \textit{(approval number: 11837)}. 
The proposed method was developed for continuous pain monitoring, aiming to enhance pain assessment protocols and improve patient care. However, as validation and testing were performed on controlled laboratory data, its deployment in real-world clinical settings requires further investigation and comprehensive evaluation.

\section*{Acknowledgements}
This paper is supported by the projects that have received funding from the
European Union's Horizon 2020 research and innovation programme under grant agreement
$101080905$ (\textit{STRATIFYHF project}).

\bibliographystyle{ACM-Reference-Format}
\bibliography{library}

\end{document}